\documentclass{article}

\usepackage[margin = 1 in]{geometry}
\usepackage{graphicx}
\usepackage{amsmath}
\usepackage{natbib}
\usepackage{authblk}

\title{Explainable Neural Networks based on Additive Index Models}
\author{Joel Vaughan\thanks{Email: \texttt{Joel.Vaughan@wellsfargo.com}; Corresponding author}
, Agus Sudjianto, Erind Brahimi, Jie Chen, and Vijayan N. Nair}
\affil{Corporate Model Risk, Wells Fargo, USA}
\date{June 2018}

\begin{document}

\maketitle

\begin{abstract}
Machine Learning algorithms are increasingly being used in recent years due to their flexibility in model fitting and increased predictive performance. However, the complexity of the models makes them hard for the data analyst to interpret the results and explain them without additional tools. This has led to much research in developing various approaches to understand the model behavior. In this paper, we present the Explainable Neural Network (xNN), a structured neural network designed especially to learn interpretable features. Unlike fully connected neural networks, the features engineered by the xNN can be extracted from the network in a relatively straightforward manner and the results displayed. With appropriate regularization, the xNN provides a parsimonious explanation of the relationship between the features and the output. We illustrate this interpretable feature--engineering property on simulated examples. 
\end{abstract}

\section{Introduction}

Neural networks (NNs) and ensemble algorithms such as Gradient Boosting Machines (GBMs) and Random Forest (RFs) have become popular in recent years due to their predictive power and flexibility in model fitting. They are especially useful with large data sets where it is difficult to do handcrafted variable selection and feature engineering. Further, in these situations, they have substantially better predictive performance compared to traditional statistical methods. Despite these advantages, there has been reluctance to fully adopt them.  One of the primary barriers to widespread adoption is the ``black box'' nature of such models.  The models are very complex and cannot be written down explicitly.   It is therefore difficult for a modeler to explain the relationships between the input features and the response or more generally understand the model's behavior. However, the ability to interpret a model and explain its behavior is critical in certain industries such as medicine and health care that deal with high risk or in banking and finance that are strongly regulated. For instance, in banking, regulators require that the input-output relationships are consistent with business knowledge and the model includes key economic variables that have to be used for stress testing.

These challenges have led to a lot of research recently in developing tools to ``open up the black box''. There are, broadly speaking, three inter-related model--based areas of research: a) global diagnostics (\cite{SobKuch2009}, \cite{Kuch2010}); b) local diagnostics ( \cite{SunTalYan2017}, \cite{AncoCeo2018}); and c) development of approximate or surrogate models that may be easier to understand and explain. These models which may be either global (\cite{HinVin2015}, \cite{CucCar2006}, \cite{TanCar2018}) or local (\cite{HuChen2018}) in nature.  There are also efforts to understand neural networks using visualization--based techniques such as those described in \cite{KahnAnd2017} or \cite{OlahMord2017}.

In this paper, we propose a flexible, yet inherently explainable, model.   More specifically, we describe a structured  network that  imposes some constraints on the network architecture and thereby provides better insights into the underlying model. We refer to it as {\it explainable neural network}, or xNN.  The  structure provides a means to understand and describe the features engineered by the network in terms of linear combinations of the input features and univariate non-linear transformations. 

\begin{itemize}
\item We use the terms ``interpretable'' and ``explainable'' interchangeably in this paper although, strictly speaking, they have different meanings. The former refers to the ability to understand and interpret the results to yourself; and the latter is the ability to explain the results to someone else. So interpretability can be viewed as a precursor to explainability. But we do not make that distinction in this paper.
\item Explainability by itself is not enough without also considering predictive performance. For instance, a linear model is very explainable but it is likely to have poor performance approximating a complex surface. In the simple examples considered in the paper, the xNNs have excellent predictive performance. But additional research is needed on more complex examples, and this is being currently pursued.
\end{itemize}

Feedforward neural networks typically consist of fully connected layers, i.e.,  the output of each node on layer $i$ is used as input for each node on layer $i+1$.  By limiting the connections between nodes, we can give a feedforward neural network  structure that can be exploited for different purposes. For example,  \citet{TsangCheng2018} considered a structure to detect interactions among input features in the presence of features' main effects.  In this paper, we propose a structured neural network designed to be explainable, meaning that it is relatively easy to describe the features and nonlinear transformations learned by the network via the network structure.   It is based on the concept of additive index models (\cite{RuanYuan2010}, \cite{Yuan11} ) and is related to projection pursuit and generalized additive models (\cite{HastTib86}).

The remainder of the paper is as follows.  In Section \ref{sec:fimxnn}, we review additive index model and introduce the explainable neural network architecture.  In Section \ref{sec:viz}, we illustrate how the components of the xNN may be used to describe the engineered features of the input variables the network learns.  Section \ref{sec:prac} discusses several practical considerations that arise in using such networks in practice.  Finally, we provide additional examples of trained xNN models in Section \ref{sec:exam}.

\section{Additive Index Models}
\label{sec:fimxnn}

\label{subsec:fim}
The formal definition of a {\bf additive index model} is given  in \eqref{eq:fim}:  

\begin{equation}
\label{eq:fim}
f(\mathbf{x}) = g_1\left(\beta_1^T \mathbf{x} \right) + g_2\left(\beta_2^T \mathbf{x} \right) 
+ \dots + g_K\left(\beta_K^T \mathbf{x} \right),
\end{equation}
where the function on the LHS can be expressed as a sum of $K$ smooth functions $g_i(\cdot)$ \citep{RuanYuan2010}. These univariate functions are each applied to a linear combination of the input features ($\beta_i^T x$).  The coefficients $\beta_i$ are often referred to as projection indices and the  $g_i(\cdot)$'s are referred to as ridge functions, following \cite{FriedStuetz81}. See also \citet{HastTib86} for the related notion of generalized additive models. The additive index model in  \eqref{eq:fim}  provides a flexible framework for approximating complex functions.  In fact, as shown in \citet{DiaShah84}, the additive index models can approximate any multivariate function $f(\mathbf{x})$ with arbitrary accuracy provided $K$, the number of ridge functions, is sufficiently large.  In practice, additive index models can be fit using penalized least squares methods to simultaneously fit the model and select the appropriate number of ridge functions  (\cite{RuanYuan2010}). See also \citet{Yuan11} for a discussion of identifiability issues surrounding such models.

\section{Explainable Neural Network Architecture (xNN)}
\label{subsec:xnn}

The Explainable Neural Network provides an alternative formulation of the additive index model as a structured neural network. It also provides a direct approach for fitting the model via gradient-based training methods for neural networks. The resulting model has built-in interpretation mechanisms as well as automated feature engineering.  We discuss these mechanisms in more detail in Section \ref{sec:viz}. Here, we describe the architecture of the xNN.

We define a modified version of the additive index model in \eqref{eq:fim} as follows: 

\begin{equation}
\label{eq:xnn}
f(\mathbf{x}) = \mu + \gamma_1h_1\left(\beta_1^T \mathbf{x} \right) + \gamma_2h_2\left(\beta_1^T \mathbf{x} \right) 
+ \dots + \gamma_K h_K\left(\beta_K^T \mathbf{x} \right).
\end{equation}
Although the shift parameter $\mu$ and the scale parameters $\gamma_k$'s are not identifiable, they are useful for the purposes model fitting: selecting an appropriate number of ridge functions through regularization.  

\begin{figure}
\centering
\includegraphics[width = 0.8\textwidth]{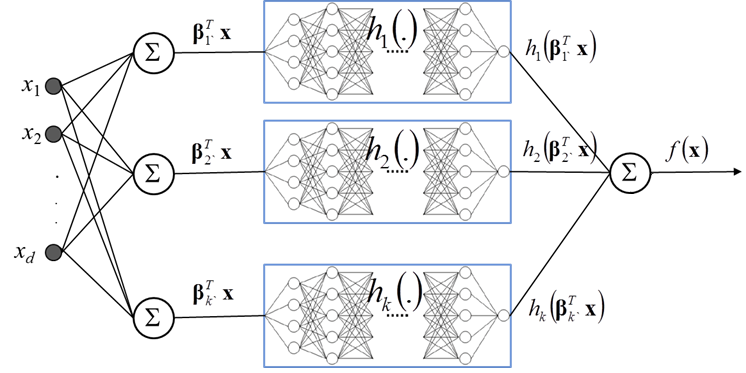}
\caption{\label{fig:xnn_arch} The xNN architecture. The three important structural components are (i) The projection layer (first hidden layer) using linear activation function.  Each node on this layer feeds into one (ii) subnetwork, which learns a potentially nonlinear transformation of the input, and the (iii) combination layer calculates a weighted sum the output of the ridge functions. }
\end{figure}

The structure of an xNN is designed to explicitly learn the model given in equation \eqref{eq:xnn}.  Figure \ref{fig:xnn_arch} illustrates the architecture of an xNN. The input layer is fully connected to the first hidden layer (called the {\it projection layer}), which consists of $K$ nodes (one for each ridge function.).  The weights of the node $i$ in the first hidden layer corresponds to the coefficients ($\beta_i$) of the input to the corresponding ridge function.  The  projection layer uses a linear activation function, to ensure that each node in this layer learns a linear combination of the input features.  The output of each node in the projection layer is used as the input to exactly one {\it subnetwork}.  

Subnetworks are used to learn the ridge functions, $h_i(\cdot)$.  The external structure of the subnetworks is essential to the xNN.  Each subnetwork must have univariate input and output, and there must be no connections between subnetworks.  The internal structure of subnetworks is less critical, provided that the subnetworks have sufficient structure to learn a broad class of univariate functions. Subnetworks typically consist of multiple fully-connected layers and use nonlinear activation functions.  More details are discussed in Section \ref{subsec:sn_struct}.

The {\it combination} layer is the final hidden layer of the xNN, and consists of a single node.  The  inputs of the node are the  univariate activations of all of the subnetworks.  The weights learned correspond to the $\gamma_i$'s in equation \eqref{eq:xnn}, and provide a final weighting of the ridge functions.  A linear activation function is used on this layer, so the output of the network as a whole is a linear combination of the ridge functions. (Note: A non--linear activation function may easily be used on the combination layer instead of a linear activation.  This changes to the formulation given in \eqref{eq:xnn} by wrapping the LHS in a further link function, as with generalized linear models.  We do not explore this generalization in detail here.)

The neural network based formulation of the additive index model provides some advantages over the traditional approach in the statistics literature.  First, it may be trained using the same mini--batch gradient--based methods, allowing the xNN formulation to easily be trained on datasets that may be too large to fit in memory at the same time.  Further, the neural network formulation allows the xNN to take advantage of the advancements in GPU computing used to train neural networks in general.  Finally, the neural network formulation allows for straightforward computation of partial derivatives of the function learned by the xNN.  This supports the ability to carryout derivative--based analysis techniques using the xNN, without needing to rely on finite difference approximations and the difficulties that these may cause.  Some techniques that may be employed are presented in \cite{SobKuch2009} and \cite{Kuch2010}. 

In the next section, we illustrate how the structures built into the xNN, namely the projection layer and subnetworks, provide a mechanism to explain the function learned by such a network.

\section{Visualization and Explainability of the xNN} 
\label{sec:viz}
\begin{figure}
\centering
\includegraphics[width = 0.6\textwidth]{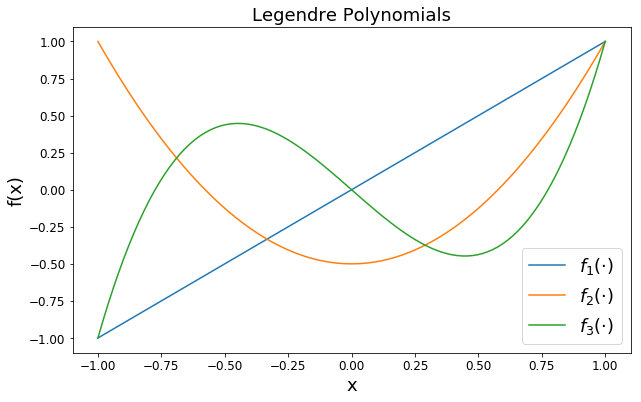}
\caption{\label{fig:legendre} The first three Legendre polynomials.}
\end{figure}

We now illustrate how visualization of xNN components can be used to aid in explainability.  We consider a simple toy example based on the first three Legendre polynomials, shown in Figure \ref{fig:legendre} and defined in \eqref{eq:legendre}.  These polynomials are orthogonal on the the interval $[-1, 1]$ and have a range of $[-1, 1]$ over the same interval.   The exact form of these functions is not of particular interest  except for the fact that they provide distinct linear, quadratic, and cubic functions on a similar scale and are orthogonal. 

\begin{equation}
\label{eq:legendre}
f_1(x) = x; \quad \quad f_2(x) = \frac{1}{2}\left( 3x^2 -1 \right); \quad \quad 
f_3(x) = \frac{1}{2}\left(5 x^3 - 3 x \right)
\end{equation}

We simulated five independent variables, $x_1, \dots, x_5$ from a Uniform distribution on $[-1, 1]$.  We then generated $y$ via

\begin{equation}
y = f_1(x_1) + f_2(x_2) + f_3(x_3) 
\end{equation}
where $f_i(\cdot), i=1,2,3$ are the Legendre polynomials as described in \eqref{eq:legendre}. This leaves $x_4, x_5$ as noise variables.  

We then built an xNN model  with 5 subnetworks and them on all five features ($x_1, \dots, x_5$).  Only the strength of the $\ell 1$ penalty on the projection and output layers were tuned.  The resulting xNN was used to generate the summaries that follow.

\subsection{Visualizing Ridge Functions}
\label{sec:vridge}
\begin{figure}
\centering
\includegraphics[width = 0.7 \textwidth]{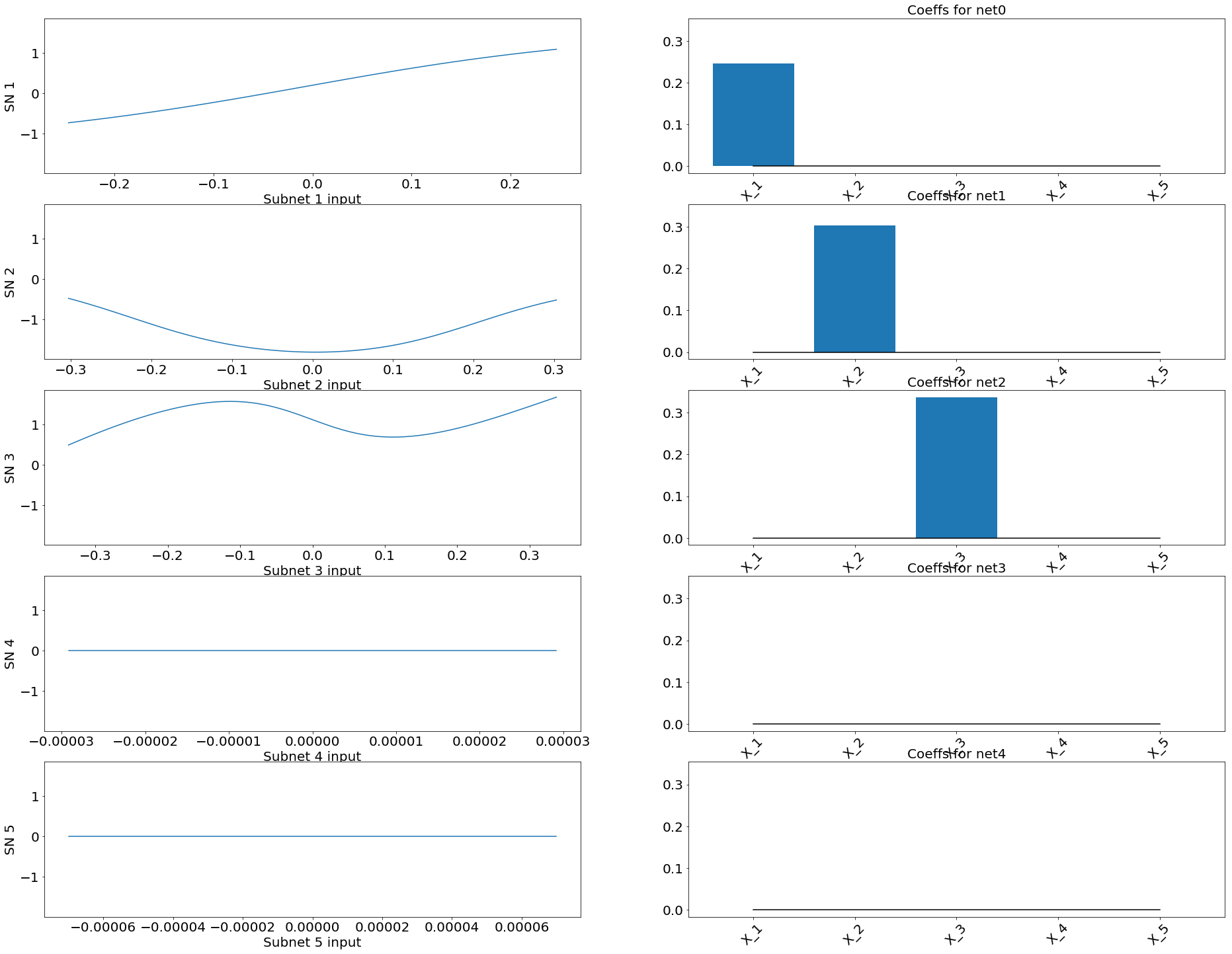}
\caption{\label{fig:subnet_toy}The ridge functions (left) and corresponding projection indices (right) of a trained xNN.}
\end{figure}

Figure~\ref{fig:subnet_toy} shows the ridge functions.  Row $i$ represents subnetwork $i$ for $i = 1, \dots, 5$.  The first column illustrates the univariate functions learned by subnetwork $i$, scaled by $\gamma_i$ .  These plots illustrate the univariate, non--linear transformations learned by the xNN in training.  The second column displays the values of $\beta_i$, the projection coefficients.  The projection coefficients explain which combination of input features is used as input to each of the ridge functions.  In this way, the plot displays the most relevant features of the network: the scaled ridge functions and the projection coefficients.

In this example, we see from Figure~\ref{fig:subnet_toy} and \ref{fig:x_toy} that Subnetwork 1 has learned the cubic Legendre function ($f_3(\cdot)$), and from the second column of Figure \ref{fig:subnet_toy}, only $x_3$ has a non-zero coefficient in the input to this subnetwork.  Subnetwork 2 has learned the quadratic function ($f_2(\cdot)$), and only the coefficient of $x_2$ is nonzero.  Subnetwork 5 has learned the linear function ($f_1(\cdot)$), and only the coefficient of $x_1$ is non-zero.  The other subnetworks (3 and 4) are not needed, and are set to zero by using an $\ell 1$ penalty on the ridge function weights ($\gamma_i$ in \eqref{eq:xnn}).

\subsection{Visualizing Univariate Effects}
\label{subsec:vx}
\begin{figure}
\centering
\includegraphics[width = 0.7 \textwidth]{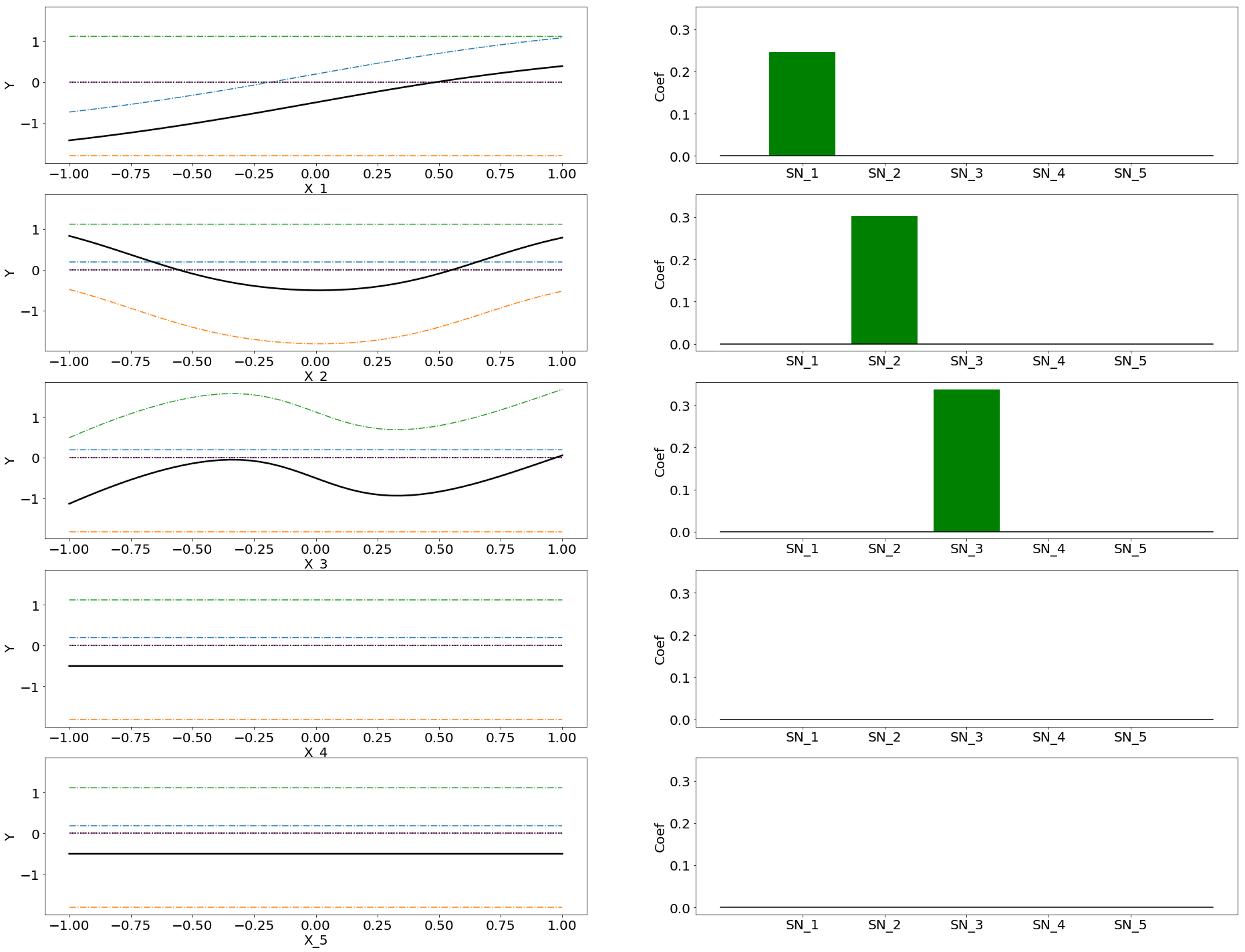}
\caption{\label{fig:x_toy} The conditional effect plots of the example xNN.}
\end{figure}

The plot shown in Figure~\ref{fig:x_toy} illustrates the feature-centric view of the xNN, which we refer to as {\it conditional effects}.  In this view, the $j$th row summarizes the xNN's treatment of the $j$th feature.  In the first column, each subnetwork's treatment of feature $j$ is plotted in row $j$, calculated via $\gamma_i h_i(\beta_{i,j} x_j)$  Each dotted line represents one such subnetwork, while the bold, solid line represents the effect of the network as a whole on feature $j$.  This is calculated via $\sum_{i=1}^k \gamma_i h_i(\beta_{i,j} x_j)$, and is the sum of the conditional effects of the individual subnetworks.  This is equivalent to plotting $f(x_i | x_j = 0), \quad \forall j \neq i$.  If the data have been standardized (as is typical in this case), this is equivalent to plotting $f(x_i | x_j = \overline x_j), \quad \forall j \neq i$.

The second column of Figure~\ref{fig:x_toy} shows the projection coefficient of feature $j$ for each of the subnetworks.  This shows which ridge functions are used to describe the effects of $x_j$.

In this particular example, we see that the only nonzero coefficient of  $x_1$ is in the projection for subnetwork 5, the linear function, and that the conditional effect on $x_1$ is linear.  Similarly, the only nonzero coefficient of $x_2$ appears in subnetwork 2, which learned a quadratic function.  The only nonzero coefficient of $x_3$ is in subnetwork 1, which has learned the cubic function ($f_3(\cdot)$).  The two extraneous variables, $x_4$ and $x_5$, have no non-zero coefficients, so the overall conditional effect of these variables is constant.

It should be mentioned that the conditional effects plot shows some information that is redundant with the subnetwork--centric view.  Nonetheless, the alternate view can be useful in understanding the role each feature plays in the predictions of the xNN model.

In this toy example, the effect of each feature is represented by exactly one ridge function.  In situations with more complex behavior, multiple ridge functions may be involved in representing the effect of a particular variable, and often are in more complex situations.  Furthermore, in under-regularized networks, the effects of each variable may be be modeled by the contributions of several subnetworks. This behavior is displayed in the examples in Section \ref{sec:exam}.

\section{Practical Considerations}
\label{sec:prac}
In this section, we consider some of the practical considerations that arise when using such models.  These include a brief discussion on the difference between model recoverability and explainability,  regularization of the xNN needed to learn a parsimonious model, and the structure of the subnetworks. 

\subsection{Model Recoverability and Explainability}
\label{subsec:expvint}

In practice, fitted xNN models exist on a spectrum of {\it model recoverability} while retaining a high degree of  {\it explainability}.  By {\it model recoverability}, we refer to the ability to recover the underlying generative mechanisms for the data., and {\it explainability} refers to  the xNN's ability to provide an explanation of the mechanisms used by the network to approximate a complex multivariate function, even if these mechanisms do not faithfully recover the underlying data generating process.  With proper regularization, as discussed in Section~\ref{subsec:regular}, the representation is parsimonious and straightforward to interpret.   The example discussed previously in Section \ref{sec:viz} illustrates a situation where the xNN has high model recoverability, meaning that it has clearly learned the underlying generating process.  In practice, this is not always be the case, as the data--generating process may not be fully described by the additive index model.    In Section \ref{subsec:nonlin}, we see such an example where the model is explainable even though it does not have higher model recoverability.

In practice, the user will never know on which end of the spectrum a given xNN sits. However, unlike other popular network structures (such as feedforward networks) or tree-based methods, the xNN has a built-in mechanism to describe the complex function learned by the network in the relatively simple terms of projections and  univariate ridge functions that ensure the model is explainable, regardless of where it may fall on the model recoverability spectrum.

Finally, note that in certain circumstances, model recoverability may not be desirable.  If the data generating process is highly complex, the explainable xNN is likely to be more easily understood given its additive nature.  The xNN is especially easy to understand if it has been properly regularized.

\subsection{Regularization and Parsimony}
\label{subsec:regular}

The overall explainability of the network can be enhanced by using an $\ell 1$ penalty on both the first and last hidden layers during training. That is, both the projection coefficients ($\beta_j$ 's) and the ridge function weights ($\gamma_i$ 's) are penalized.  When the strength of the penalty is properly tuned, this can produce a parsimonious model that is relatively easily explained.

An $\ell 1$ penalty on the first hidden layer forces the projection vectors $\beta_i$ to have few non-zero entries, meaning that each subnetwork (and corresponding ridge function) is only applied to a small set of the variables.  Similarly, an $\ell 1$ penalty on the final layer serves to force $\gamma_i$ to zero in situations where fewer subnetworks are needed in the xNN than are specified in training.

\subsection{Subnetwork Structure}
\label{subsec:sn_struct}
In principle, the subnetwork structure must be chosen so that each subnetwork is capable of learning a large class of univariate functions.  In our experience, however, both the explainability and predictive performance of the network are not highly sensitive to the subnetwork structure.  In our simulations, we have found that using subnetworks consisting of two hidden layers with structures such as [25, 10] or even [12,6] with nonlinear activation functions (tanh, e.g.) are sufficient to learn sufficiently flexible ridge functions in fitting the models.

\subsection{xNN as a Surrogate Model}
\label{subsec:surrogate}

While the xNN architecture may be used as an explainable, predictive model built directly from data, it may also be used as a surrogate model to explain other nonparametric models, such as tree-based methods and feedforward neural networks, called a {\it base model}.  Because the xNN is an explainable model, we may train an xNN using the input features and corresponding response values predicted by the base model.  We then may use the xNN to explain the relationships learned by the base model.  For further discussion of surrogate models, see  \cite{HinVin2015}, \cite{CucCar2006}, or  \cite{TanCar2018}.  The use of more easily interpretable surrogate models to help interpret a complex machine learning model is similar to the field of computer experiments, where complicated computer simulations of physical systems are studied using well--understood statistical models, as described in \citet*{FangLiSudj2005} and \cite{BastOHag2009}.  In computer experiments, the input to the computer simulation may be carefully designed to answer questions of interest using these statistical models, where as the complex ML models often restricted to observational data.

\section{Simulation Examples}
\label{sec:exam}
In this section, we illustrate the behavior of xNN networks with two simulations.  In the first, data are generated from a model that follows the additive index model framework.  This is an example where the trained xNN has high model recoverability, meaning it recovers correctly the data generating mechanism.  The second simulation does not follow the additive index model framework, yet the trained xNN is still {\it explainable}, in the sense that the xNN still provides a clear description of the mechanisms  the xNN learns to approximate the underlying response surface.

\subsection{Example 1: Linear Model with Multiplicative Interaction}
\label{subsec:interact}
We simulate six independent variables, $x_1, \dots, x_6$ from independent Uniform distributions on $[-1, 1]$.  We then generate $y$ via

\begin{equation}
\label{eq:interact}
y = 0.5x_1+0.5x_2^2+0.5x_3x_4+0.3x_5^2 + \epsilon, \quad \quad \text{ where } \epsilon\sim N(0,0.05).
\end{equation}

This is a linear model with a multiplicative interaction.  The variable, $x_6$ is left as a noise feature.  While this model does not, at first glance, fit the additive index model framework, we note that a multiplicative interaction may be represented as the sum of quadratic functions, as shown in \eqref{eq:quad}.

\begin{equation}
\label{eq:quad}
xy = c(ax+by)^2 - c(ax-by)^2  \text{ for any } a, b \text{ satisfying } ab \neq 0, a^2+b^2 = 1 \text{ with } c=1/4ab 
\end{equation}

\begin{figure}
\centering
\includegraphics[width = 0.8\textwidth]{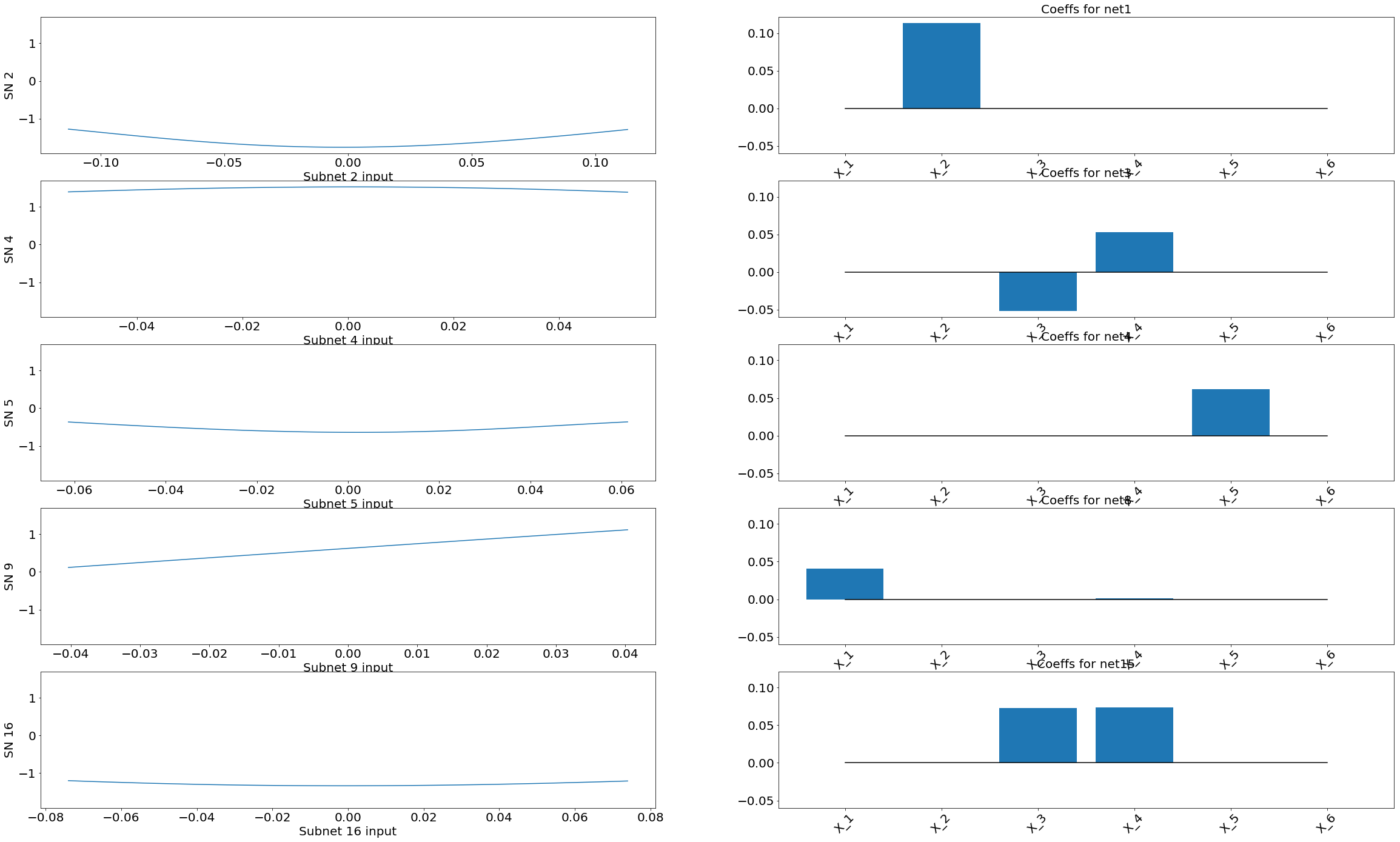}
\caption{\label{fig:subnet_interact}  The subnetwork plot of the xNN representing the linear model from Section \ref{subsec:interact}.  Only the active subnetworks are shown.}
\end{figure}

Therefore, this model may be exactly represented by an xNN.  We trained the xNN using 20 subnetworks.   The network achieved a mean squared error of 0.0028 the holdout set, close to the simulation lower bound of 0.0025. The resulting {\it active} ridge functions are illustrated in Figure \ref{fig:subnet_interact}.  (By {\it active} ridge functions, we mean those functions that are not constant.) Subnetwork 9 learned a linear ridge function, and has a relatively large projection coefficient for $x_1$.  Subnetworks 2, 4, 5, and 16 learned quadratic ridge functions.  Based on the projection coefficients, we see that subnetworks 2 and 5 are used to represent the contributions of $x_2$ and $x_5$, respectively.  Subnetworks 4 and 16 combine to represent the interaction $x_3x_4$.  Both are quadratic.  The two features have the same projection coefficients in subnetwork 16, while they have projection coefficients of opposite signs in subnetwork 4. This is exactly the representation of an interaction term described in equation \eqref{eq:quad}.  Thus, this xNN has both high model recoverability and a high degree of explainability.

\begin{figure}
\centering
\includegraphics[width =0.8\textwidth]{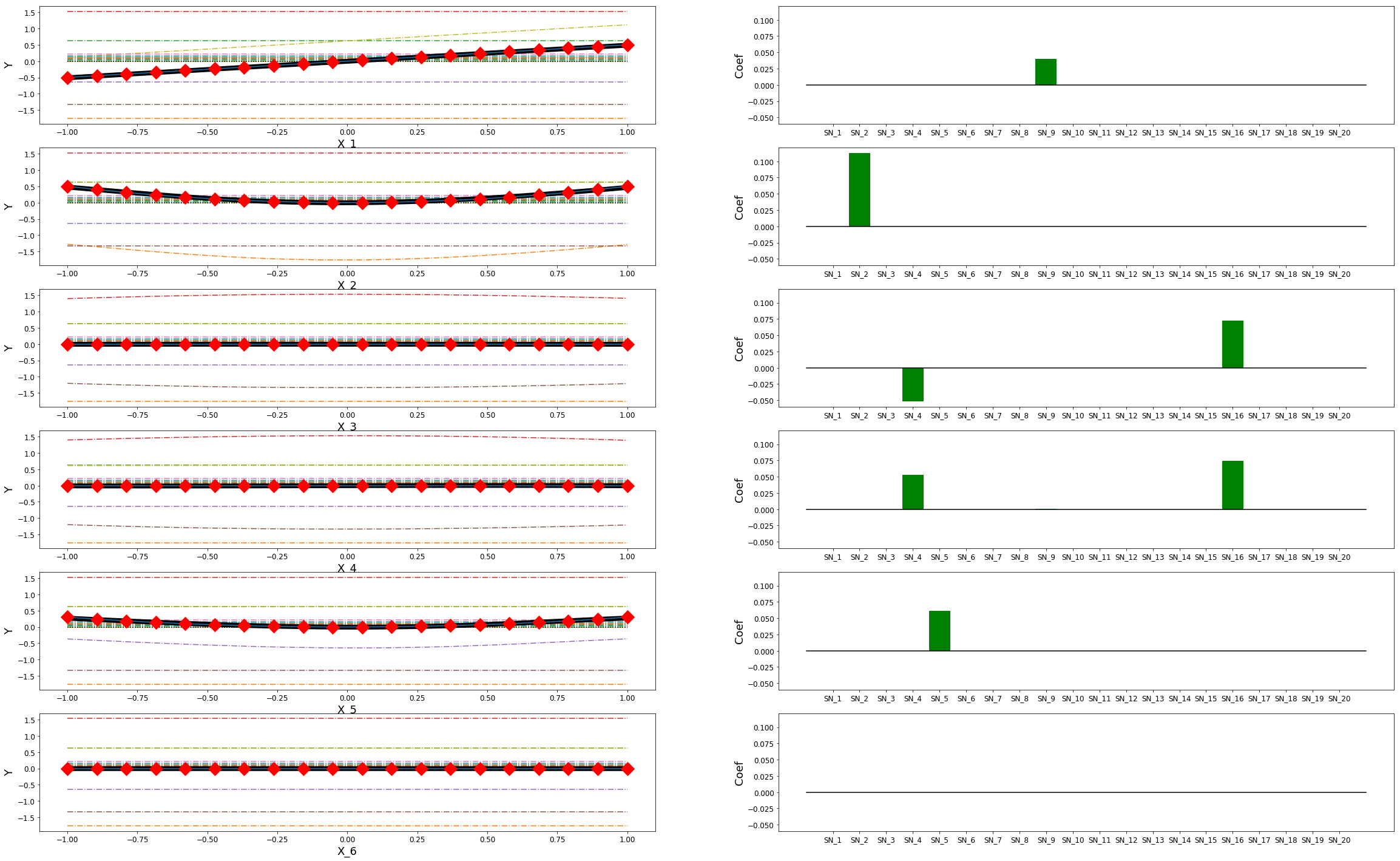}
\caption{\label{fig:x_interact} The conditional effect plots of the xNN for the linear model (Example 1).  The conditional effect plots for each feature are shown in the left column, where the thick solid black line shows the conditional effect learned by the network.  For reference, the true conditional effects (from the simulation) are shown with red markers.  Note that the estimated conditional effects match well.  The projection coefficients for the variable summarized in each row are given in the right column.  This shows the projection coefficients for $x_i$ in each of the subnetworks. }
\end{figure}

Figure~\ref{fig:x_interact} illustrates the conditional effects of each network on each of the predictors.  We see, as expected, a linear marginal effect on $x_1$ and quadratic effects on $x_2$ and $x_5$.  It is notable that the conditional effects plots for both $x_3$ and $x_4$ show no conditional effect.  In the case of such interactions, this is expected.  In this model, if we condition on e.g. $x_4=0$, then $x_3$ will show no effect on the response.  Similarly, we see no effect of $x_4$ when conditioning on $x_3=0$.

\subsection{Example 2: Non-Linear Model}
\label{subsec:nonlin}
We simulate four independent variables, $x_1, \dots, x_4$ from independent Uniform distributions on $[-1, 1]$.  We then generate $y$ via

\begin{equation}
\label{eq:nonlin}
y = \exp(x_1)\cdot\sin(x_2) + \epsilon  \quad \quad \text{ where } \epsilon\sim N(0,0.1).
\end{equation}

Both $x_3$ and $x_4$ are left as noise variables.  We then fit an xNN with 10 subnetworks and a subnet structure of [12,6] with tanh activation.  The network achieved a mean squared error of 0.0122 on a holdout test set, close to the simulation lower bound of 0.01.

Note that this generating model does not fit the additive index model framework.  In this example, the trained xNN is {\it explainable} despite having low model recoverability. Although the xNN cannot recover the data generating process, it still fits the data well, and clearly explains the mechanisms it uses to do so, by displaying the projection coefficients and learned ridge functions.

\begin{figure}
\centering
\includegraphics[width = \textwidth]{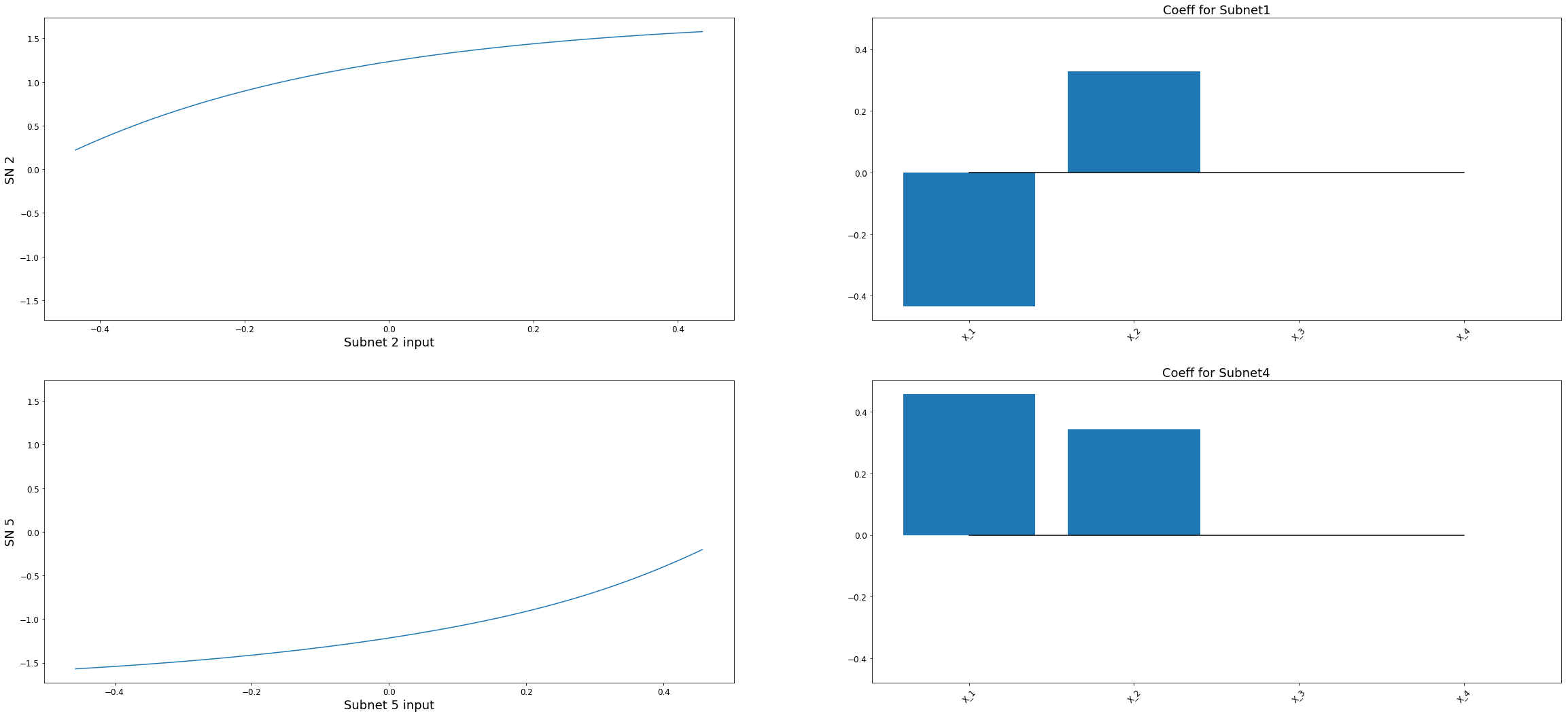}
\caption{\label{fig:subnet_nonlin} Subnetwork plot of the xNN modeling the function given by equation \eqref{eq:nonlin}.  Note that the two important variables, $x_1$ and $x_2$ are represented by two of the subnetworks.  Although not the true generating functions, the xNN is able to explain the projections and ridge functions learned to represent this model. }
\end{figure}

Figure \ref{fig:subnet_nonlin} shows two ridge functions, represented by subnetworks 2 and 5.  Both subnetworks have non-zero coefficients of $x_1$ and $x_2$, although they have the same sign in Subnetwork 5, and opposite signs in Subnetwork 2.  We see that the xNN approximates the simulated function with the function $f_2(-0.43 x_1 + 0.33 x_2) + f_5(0.46 x_1 + 0.34 x_2)$, where $f_2(\cdot)$ and  $f_5(\cdot)$ are the two ridge functions learned by subnetworks 2 and 5, respectively.

\begin{figure}
\centering
\includegraphics[width = \textwidth]{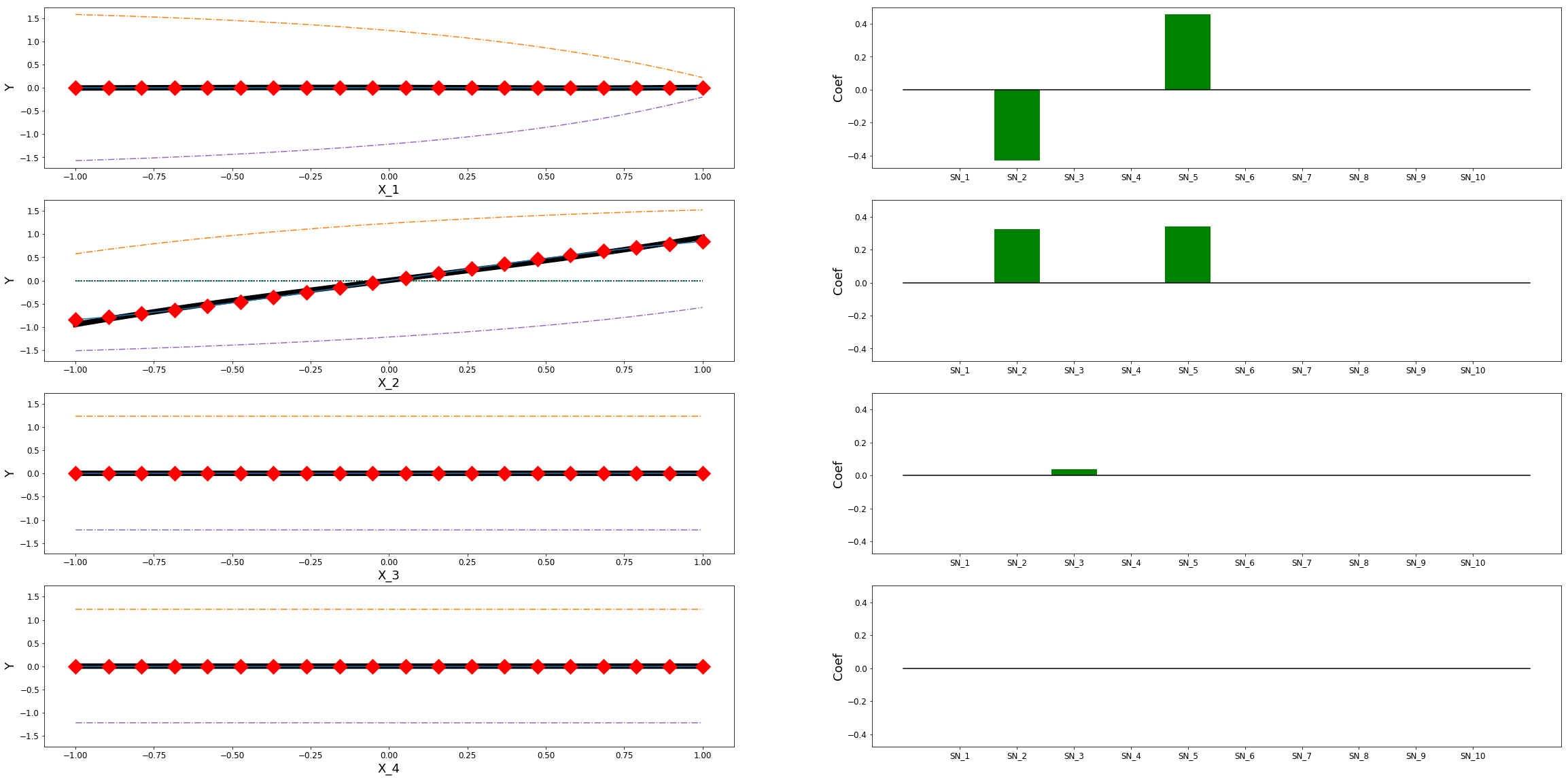}
\caption{\label{fig:x_nonlin}  Conditional Effects plot for the xNN modeling the simulation given by equation \eqref{eq:nonlin}. The conditional effect plots for each feature are shown in the left column, where the thick solid black line shows the conditional effect learned by the network.  For reference, the true conditional effects (from the simulation) are shown with red markers.  Note that the estimated conditional effects match well.  The projection coefficients for the variable summarized in each row are given in the right column.  This shows the projection coefficients for $x_i$ in each of the subnetworks.}
\end{figure}

Figure~\ref{fig:x_nonlin} shows the Note that subnetwork 3 has learned a small non-zero coefficients for $x_3$, however, the corresponding ridge function is constant at zero, so $x_3$ does not contribute to the output.  This type of behavior may occur when the xNN is slightly under regularized.

\section{Conclusion}

We have proposed an explainable neural network architecture, the xNN, based on the additive index model.  Unlike commonly used neural network structures, the structure of the xNN describes the features it learns, via linear projections and univariate functions.  These explainability features have the attractive feature of being additive in nature and straightforward to interpret. Whether the network is used as a primary model or a surrogate for a more complex model, the xNN provides straightforward explanations of how the model uses the input features to make predictions.  

Future work on the xNN will study the overall predictive performance of the xNN compared to other ML models, such as GBM and unconstrained FFNNs.  We will also study the predictive performance lost when using the xNN as a surrogate model for more complex models.

\bibliographystyle{agsm}
\bibliography{refs}

\end{document}